\documentclass[conference]{IEEEtran}
\IEEEoverridecommandlockouts
\usepackage{amsmath,graphicx}
\usepackage{amsfonts}
\usepackage{amsmath}
\usepackage{bm,amssymb,amsfonts}
\usepackage{tabularx}
\usepackage{tabulary}
\usepackage{multirow}
\usepackage{dblfloatfix}
\usepackage[dvipsnames]{xcolor}
\usepackage[inline, shortlabels]{enumitem}
\usepackage{xcolor, soul}
\usepackage{caption,color}
\usepackage{cite}
\usepackage{amsmath,amssymb,amsfonts}
\usepackage{algorithmic}
\usepackage{graphicx}
\usepackage{textcomp}
\usepackage{xcolor}
\def\BibTeX{{\rm B\kern-.05em{\sc i\kern-.025em b}\kern-.08em
    T\kern-.1667em\lower.7ex\hbox{E}\kern-.125emX}}
    
\begin{document}

\title{\textsf{SFC-GAN}: A Generative Adversarial Network for Brain Functional and Structural Connectome Translation\\
}

\author{\IEEEauthorblockN{Yee-Fan Tan}
\IEEEauthorblockA{\textit{School of IT} \\
\textit{Monash University Malaysia}\\
Subang Jaya, Malaysia \\
tan.yeefan@monash.edu}
\and
\IEEEauthorblockN{Jun Lin Liow}
\IEEEauthorblockA{\textit{School of IT} \\
\textit{Monash University Malaysia}\\
Subang Jaya, Malaysia \\
junlin.liow7@gmail.com}
\and
\IEEEauthorblockN{Pei-Sze Tan}
\IEEEauthorblockA{\textit{School of IT} \\
\textit{Monash University Malaysia}\\
Subang Jaya, Malaysia \\
tan.peisze@monash.edu}
\and
\IEEEauthorblockN{Fuad Noman}
\IEEEauthorblockA{\textit{School of IT} \\
\textit{Monash University Malaysia}\\
Subang Jaya, Malaysia \\
fuad.noman@monash.edu}
\and
\IEEEauthorblockN{Rapha\"{e}l C.-W. Phan}
\IEEEauthorblockA{\textit{School of IT} \\
\textit{Monash University Malaysia}\\
Subang Jaya, Malaysia \\
raphael.phan@monash.edu}
\and
\IEEEauthorblockN{Hernando Ombao}
\IEEEauthorblockA{\textit{Statistics Program} \\
\textit{King Abdullah University of Science and Technology}\\
Thuwal, Saudi Arabia \\
hernando.ombao@kaust.edu.sa}
\and
\IEEEauthorblockN{Chee-Ming Ting}
\IEEEauthorblockA{\textit{School of IT} \\
\textit{Monash University Malaysia}\\
Subang Jaya, Malaysia \\
ting.cheeming@monash.edu}
}

\maketitle

\begin{abstract}
Modern brain imaging technologies have enabled the detailed reconstruction of human brain connectomes, capturing structural connectivity (SC) from diffusion MRI and functional connectivity (FC) from functional MRI. Understanding the intricate relationships between SC and FC is vital for gaining deeper insights into the brain's functional and organizational mechanisms. However, obtaining both SC and FC modalities simultaneously remains challenging, hindering comprehensive analyses. Existing deep generative models typically focus on synthesizing a single modality or unidirectional translation between FC and SC, thereby missing the potential benefits of bi-directional translation, especially in scenarios where only one connectome is available. Therefore, we propose Structural-Functional Connectivity GAN (\textsf{SFC-GAN}), a novel framework for bidirectional translation between SC and FC. This approach leverages the CycleGAN architecture, incorporating convolutional layers to effectively capture the spatial structures of brain connectomes. To preserve the topological integrity of these connectomes, we employ a structure-preserving loss that guides the model in capturing both global and local connectome patterns while maintaining symmetry. Our framework demonstrates superior performance in translating between SC and FC, outperforming baseline models in similarity and graph property evaluations compared to ground truth data, each translated modality can be effectively utilized for downstream classification.

\end{abstract}


\vspace{-0.2cm}
\section{Introduction}
\label{sec:intro}

Advancements in brain imaging and tracking technologies have enabled detailed reconstruction of human brain connectomes, where nodes represent distinct brain regions and network edges capture either structural connectivity (SC) or functional connectivity (FC). SC corresponds to white matter pathways inferred from diffusion magnetic resonance imaging (dMRI), while FC reflects statistical relationships derived from functional signals like functional magnetic resonance imaging (fMRI) \cite{chu2018function}. Understanding the intricate relationship between SC and FC is essential for gaining deeper insights into the brain's functional and organizational mechanisms and how these might be disrupted in neurological disorders \cite{wein2021brain, zamani2022local}. Prior studies have shown a strong correlation between SC and FC \cite{skudlarski2008measuring}, and both modalities provide complementary information, which can significantly improve the accuracy of brain disorder diagnosis \cite{wee2012identification, li2024machine, tan2023unified, zuo2023alzheimer}. However, obtaining both SC and FC data simultaneously is often challenging.

Deep learning models have shown significant promise in synthesizing high quality brain network data, such as generating brain FC \cite{li2021brainnetgan,yao2019brain,barile2021data, tan2024fmri, tan2023unified, tan2024brainfc} or SC \cite{li2021brainnetgan,barile2021data,kong2022adversarial}, for downstream classification tasks. Nonetheless, these approaches typically focus on a single modality of connectomes, failing to address the challenges posed by the absence of both SC and FC.

To tackle the above issue, recent works have leveraged graph-based neural networks in relating human brain structure-function relationships by predicting SC from FC \cite{zhang2022predicting,zhang2020recovering} or FC from SC \cite{neudorf2022structure, li2022learning, li2020supervised}. However, these studies are limited by their focus on unidirectional prediction, neglecting the potential benefits of bi-directional SC-FC translation, which allows inverse mapping of the connectomes.



\begin{figure*}[!htp]
\captionsetup{skip=-3pt}
\centering
\includegraphics[width=0.9\textwidth]{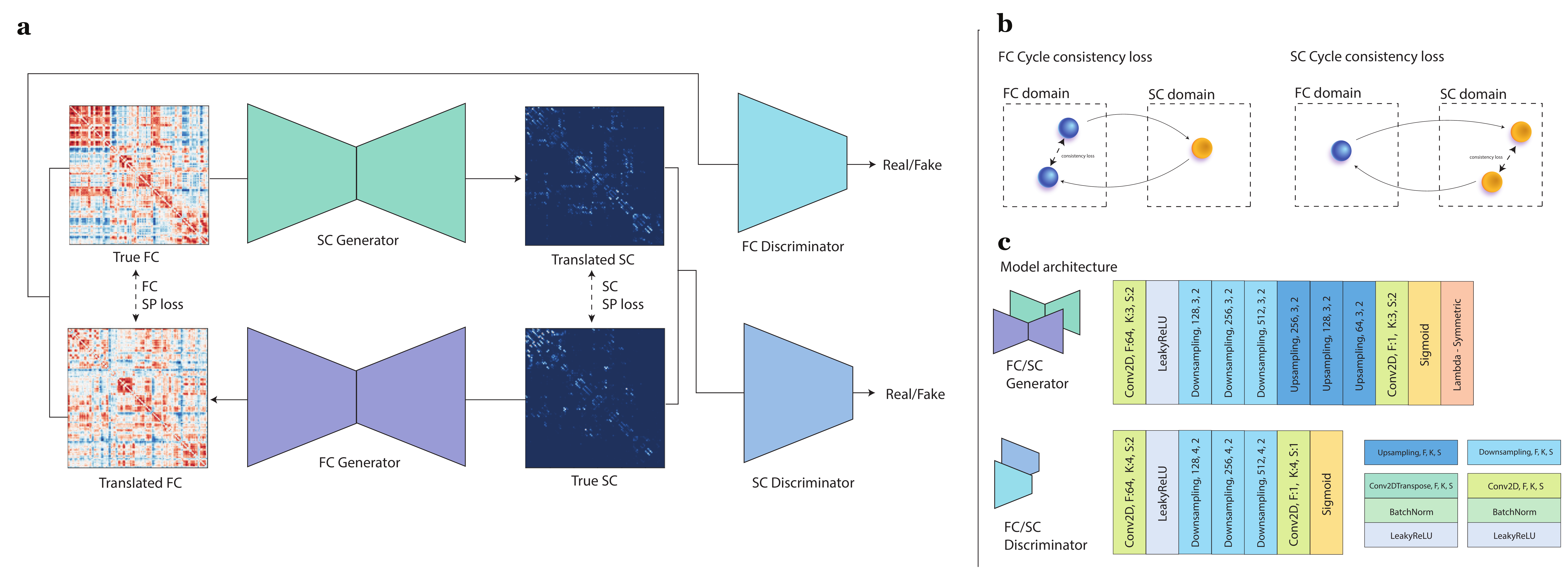}
\caption{Overview of \textsf{\textsf{SFC-GAN}}: \textbf{a} Training of \textsf{SFC-GAN}, $G_{SC}$ translates FC, $\bm{x}_{fc}$ to SC $\tilde{\bm{x}}_{sc}$, where $G_{FC}$ translates SC, $\bm{x}_{sc}$ to FC $\tilde{\bm{x}}_{fc}$. $D_{FC}$ and $D_{SC}$ aim to discriminate between $\bm{x}_{fc}$ and $\tilde{\bm{x}}_{fc}$, $x_{sc}$ and $\tilde{\bm{x}}_{sc}$, respectively. \textbf{b} Cycle consistency loss of FC and SC domains. \textbf{c} Network architectures of $G_{FC}$, $G_{SC}$, $D_{FC}$, and $D_{SC}$.
}
\label{fig:overview}
\vspace{-0.5cm}
\end{figure*}

Motivated by the aforementioned challenges, we propose Structural-Functional Connectivity Generative Adversarial Network (\textsf{SFC-GAN}), a novel framework enabling direct and inverse mapping of individual SC and FC. The key contributions of this work are as follows: \begin{enumerate*} \item[(1)] To the best of our knowledge, this is the first study to explore bidirectional translation between structural and functional connectomes. We leverage the CycleGAN architecture \cite{zhu2017unpaired}, incorporating convolutional layers to effectively capture the intricate spatial structure of brain connectomes. \item[(2)] To maintain the topological integrity of the connectomes, we employ a structure-preserving loss inspired by \cite{zhang2022predicting}, which guides the model in capturing both global and local connectome patterns. Additionally, we design the network architecture to preserve the symmetry inherent in these connectomes. \textsf{SFC-GAN} demonstrates superior performance in translating connectomes, and unveils individual structure-function relationships, making it particularly valuable in scenarios where one of the connectomes is unavailable.
\end{enumerate*}

\section{Methods}
\label{sec:methods}

\subsection{Problem Formulation}

We estimate brain FC as an $n \times n$ cross-correlation matrix between fMRI time series and SC as the fiber counts over $n$ brain regions of interest, respectively. 
Let $\mathcal{X_{\text{FC}}}$ and $\mathcal{X_{\text{SC}}}$ be a set of FCs and SCs, such that $\bm{x}_{\text{FC}} \in \mathcal{X_{\text{FC}}}$ and $\bm{x}_{\text{SC}} \in \mathcal{X_{\text{SC}}}$. Given $\bm{x}_{\text{FC}}$, we aim to translate it to SC $\tilde{\bm{x}}_{\text{SC}}$, and vice versa from $\bm{x}_{\text{SC}}$ to $\tilde{\bm{x}}_{\text{FC}}$, while preserving the connectome topology. The ultimate goal is to learn mapping functions $\theta_{\text{SC}}$ from $\bm{x}_{\text{\text{FC}}}$ to $\tilde{\bm{x}}_{\text{SC}}$, and $\theta_{\text{FC}}$ from $\bm{x}_{\text{SC}}$ to $\tilde{\bm{x}}_{\text{FC}}$.

\subsection{\textsf{SFC-GAN}}

We introduce \textsf{SFC-GAN}, an extension of the CycleGAN \cite{zhu2017unpaired}, designed to translate between individual FC and SC pairs while preserving connectome topology. The proposed model comprises two generators, $G_{\text{SC}}$ and $G_{\text{FC}}$, and two discriminators, $D_{\text{SC}}$ and $D_{\text{FC}}$ as illustrated in Fig.~\ref{fig:overview}a. 
 
Similar to CycleGAN, we adopt the adversarial loss, with $G_{\text{FC}}: X_{SC} \rightarrow \tilde{X}_{\text{FC}}$ and $G_{\text{SC}}: X_{\text{FC}} \rightarrow \tilde{X}_{\text{SC}}$, where the objective for $G_{\text{FC}}$ is expressed as:

\vspace{-0.15cm}
\begin{equation}
\begin{aligned}
\mathcal{L}_{GAN}(G_{\text{FC}}, D_{\text{SC}}, & {x}_{\text{FC}}, \tilde{x}_{\text{SC}})  = \mathbb{E}_{x_{\text{SC}} \sim p_{\text{\text{SC}}}} \left[ log  D_{\text{SC}}(x_{\text{SC}}) \right] \\  &+  \mathbb{E}_{x_{\text{FC}} \sim p_{\text{\text{FC}}}} \left[ log (1- D_{\text{SC}}(\tilde{x}_{\text{SC}}) \right]
\end{aligned}
\end{equation}

\begin{table*}[b]
\caption{Translated FC and SC similarity and graph property evaluation with the incorporation of $\mathcal{L}_{SP}$.} 
\label{tab:results_similarity}

\resizebox{\textwidth}{!}{
\begin{tabular}{cc|c|ccccc|cccc}
\hline
\multirow{2}{*}{\centering \textbf{Loss}} & \multirow{2}{*}{\centering \textbf{Dataset}} &\multirow{2}{*}{\centering \textbf{Data type}} & \multicolumn{5}{c|}{\centering \textbf{Similarity Evaluation}} & \multicolumn{4}{c}{\centering \textbf{Graph Property APD}} \\
& & & MSE & MAE & SSIM & Pearson & Cosine & Density & CPL & Efficiency &Modularity\\
\hline
\multirow{4}{*}{\centering Without $\mathcal{L}_{SP}$ } & 
\multirow{2}{*}{\centering \textbf{ADNI}} & Translated FC& $0.0230\pm0.0046$ & $0.1222\pm0.0130$ & $32.61\pm5.41$ & $34.19\pm8.76$ & $97.62\pm0.43$
									& $10.92 \pm 8.05$ & $6.65 \pm 4.97$ & $4.69 \pm 3.44$ & $75.85 \pm 37.30$\\
							   & & Translated SC & $0.0050\pm0.0001$ & $0.0214\pm0.0011$ & $31.79\pm2.57$ & $4.03\pm1.85$ & $8.94\pm0.02$
							   		& $155.79 \pm 9.38$ & $81.35 \pm 8.56$ & $70.27 \pm 8.39$ & $192.54 \pm 5.89$\\ 

\cline{2-12}
&  \multirow{2}{*}{\centering \textbf{DUMC-MDD}} & Translated FC & $0.0210\pm0.0015$ & $0.1141\pm0.0046$ & $14.76\pm1.51$ & $34.46\pm2.05$ & $95.13\pm0.31$
									& $26.35 \pm 8.41$ & $5.25 \pm 2.23$ & $6.22 \pm 2.33$ & $104.11 \pm 28.72$ \\
							  & & Translated SC & $0.0110\pm0.0001$ & $0.0305\pm0.0002$ & $37.65\pm6.84$ & $57.60\pm0.35$ & $59.20\pm3.41$
									& $74.85 \pm 10.97$ & $44.76 \pm 4.19$ & $32.68 \pm 4.24$ & $109.63 \pm 13.89$\\ 
\hline
\multirow{4}{*}{\centering With $\mathcal{L}_{SP}$ } & 
\multirow{2}{*}{\centering \textbf{ADNI}} & Translated FC& $0.0191\pm0.0055$ & $0.1099\pm0.0154$ & $37.00\pm7.18$ & $43.30\pm7.35$ & $98.03\pm5.11$
										& $8.55 \pm 6.89$ & $6.18 \pm 4.72$ & $3.89 \pm 3.09$ & $72.81 \pm 47.73$\\ 

							   & & Translated SC & $0.0001\pm0.0001$ & $0.0017\pm0.0012$ & $92.58\pm6.45$ & $55.63\pm9.20$ & $56.85\pm9.01$
							   		& $153.04 \pm 9.88$ & $77.25 \pm 8.73$ & $68.04 \pm 8.47$ & $192.39 \pm 5.97$\\

\cline{2-12}
&  \multirow{2}{*}{\centering \textbf{DUMC-MDD}} & Translated FC& $0.0237\pm0.0013$ & $0.1224\pm0.0036$ & $24.45\pm2.22$ & $41.85\pm2.06$ & $94.59\pm2.87$
									& $8.49 \pm 5.58$ & $3.99 \pm 2.02$ & $2.95 \pm 1.83$ & $30.45 \pm 25.59$\\

							&    & Translated SC & $0.0191\pm0.0055$ & $0.0083\pm0.0002$ & $81.24\pm0.67$ & $94.33\pm2.48$ & $94.49\pm2.40$
							   		& $72.26\pm11.53$ & $40.10\pm4.23$ & $30.22\pm4.25$ & $112.00\pm14.73$ \\
\hline
\hline

\end{tabular}
}
\end{table*}

\begin{figure*}[b]
\centering
\vspace{-0.5cm}
\captionsetup{skip=0pt}
\includegraphics[width=0.85\textwidth]{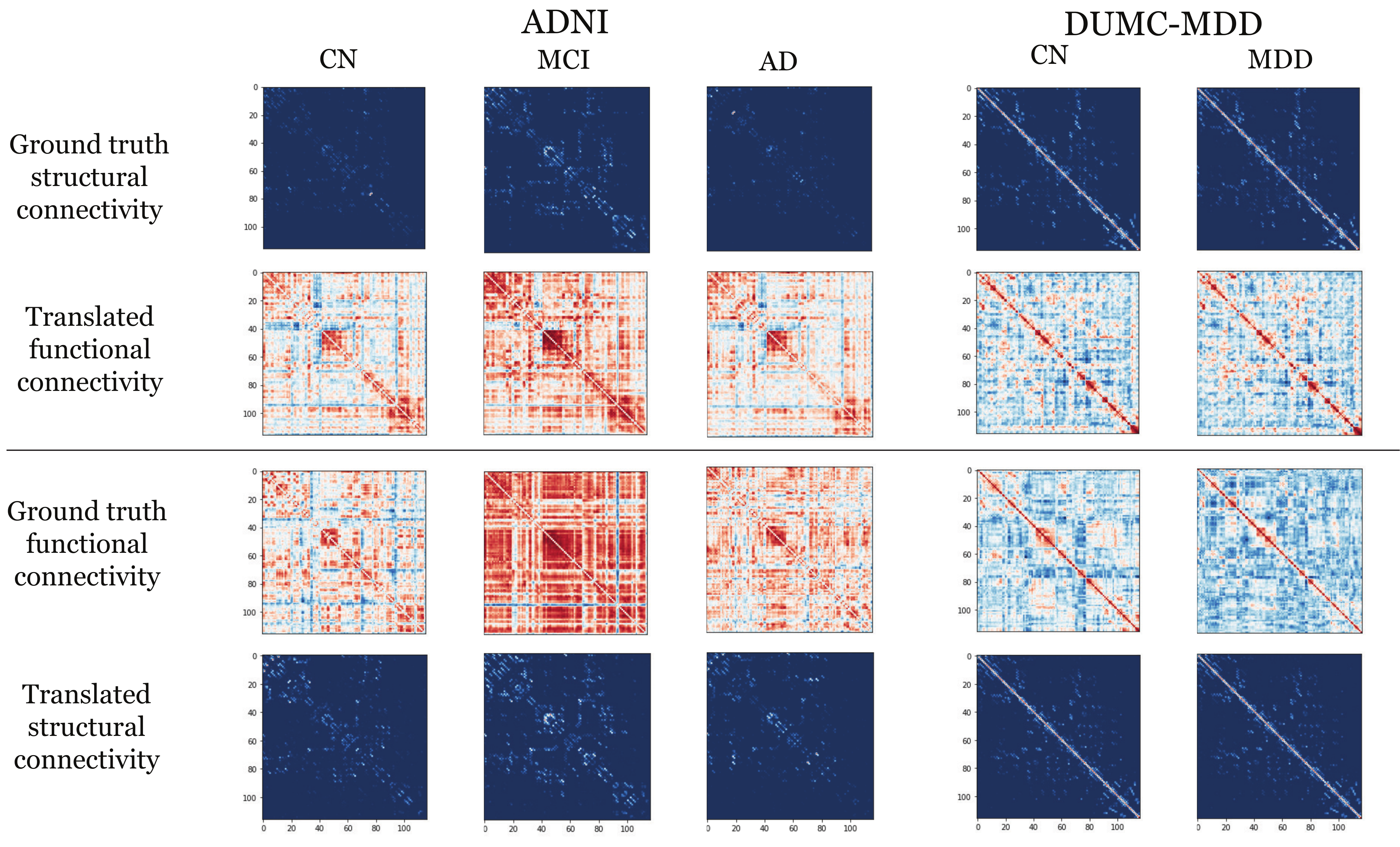}
\caption{Matrix reconstruction results on both ADNI and DUMC-MDD datasets using \textsf{SFC-GAN}.
}
\label{fig:matrix_recon}

\end{figure*}
\vspace{-0.15cm}
\noindent The loss for $G_{\text{SC}}$ follows $\mathcal{L}_{GAN}(G_{\text{SC}}, D_{\text{FC}}, {x}_{\text{SC}}, \tilde{x}_{\text{FC}})$, where $\tilde x_{\text{FC}} = G_{\text{FC}}(x_{\text{SC}})$ and $\tilde x_{\text{SC}} = G_{\text{SC}}(x_{\text{FC}})$. Consistency loss is employed to consistency mappings between FC/SC pairs (Fig.~\ref{fig:overview}b), following:

\begin{equation}
\begin{aligned}
\mathcal{L}_{cyc}(G_{\text{FC}}, G_{\text{SC}}) & = \mathbb{E}_{x_{\text{SC}} \sim p_{\text{SC}}} \left[\|\tilde{x}_{\text{SC}} - x_{\text{SC}}\|_1\right] \\
& + \mathbb{E}_{{x_{\text{FC}} \sim p_{\text{FC}}}} \left[\|\tilde{x}_{SC} - x_{\text{FC}}\|_1\right]
\end{aligned}
\end{equation}

\noindent To encourage the generators to preserve the connectome structure if the connectome belongs to the target domain, we incorporate the identity loss as in (\ref{eq:id}).


\vspace{-0.1cm}
\begin{equation}
\begin{aligned}
\mathcal{L}_{\text{id}}(G_{\text{FC}}, G_{\text{SC}}) &=\mathbb{E}_{{x_{\text{FC}} \sim p_{\text{FC}}}} \left[\| G_{\text{FC}}({x}_{FC}) - x_{\text{FC}}\|_1\right] \\
& +  \mathbb{E}_{x_{\text{SC}} \sim p_{\text{SC}}} \left[\|G_{\text{SC}}({x}_{SC}) - x_{\text{SC}}\|_1\right]
\end{aligned}
\label{eq:id}
\end{equation}

\noindent where the $L1$ norm above encourages that the generators produce identical connectomes given connectomes in the same domain, penalizing deviations. Inspired by \cite{zhang2022predicting}, we adopt the structure-preserving loss, composed of the mean squared error (MSE) and pearson correlation coefficient (PCC) loss: 
\begin{equation}
\begin{aligned}
\mathcal{L}_{SP}(G_{\text{FC}}, G_{\text{SC}}) &= \mathcal{L}_{PCC}({x_{\text{FC}}}, \tilde{x}_{SC} )  +  \mathcal{L}_{PCC}({x_{\text{SC}}}, \tilde{x}_{FC})  \\
& +  \mathcal{L}_{MSE}({x_{\text{FC}}}, \tilde{x}_{FC})   +  \mathcal{L}_{MSE}({x_{\text{SC}}}, \tilde{x}_{SC}) 
\end{aligned}
\end{equation}

\noindent where $\mathcal{L}_{MSE}(\cdot, \cdot)$ ensures element-wise similarity between connectomes, and $\mathcal{L}_{PCC}(\cdot, \cdot)$ measures the correlation between any given pairs of connectomes as defined in (\ref{eqn:pcc}). Specifically, $\mathcal{L}_{PCC-b}(\cdot, \cdot)$ and $\mathcal{L}_{PCC-r}(\cdot, \cdot)$ measure the correlations across the entire brain and within individual brain regions (each row/column of the connectome), respectively.

\begin{equation}
\begin{aligned}
&\mathcal{L}_{PCC}({x}, \tilde{{x}}) = \mathcal{L}_{PCC-b}({x}, \tilde{{x}})  + \mathcal{L}_{PCC-r}({{x}}, \tilde{{x}})  \\
&= \frac{\sum_{i=1}^n \sum_{j=1}^n \left({x}_{i,j} - \bar{x}\right) \left(\tilde{{x}}_{i,j} - \overline{\tilde{{x}}}\right)}{\sqrt{\sum_{i=1}^n \sum_{j=1}^n \left({x}_{i,j} - \bar{{x}}\right)^2} \sqrt{\sum_{i=1}^n \sum_{j=1}^n \left(\tilde{{x}}_{i,j} - \overline{\tilde{{x}}}\right)^2}} \\
&+ \sum_{i=1}^n \frac{\sum_{j=1}^n \left({x}_{i,j} - \bar{{x}}\right) \left(\tilde{{x}}_{i,j} - \overline{\tilde{{x}}}\right)}{\sqrt{\sum_{j=1}^n \left({x}_{i,j} - \bar{{x}}\right)^2} \sqrt{\sum_{j=1}^n \left(\tilde{{x}}_{i,j} - \overline{\tilde{{x}}}\right)^2}} \\
\end{aligned}
\label{eqn:pcc}
\end{equation}


\noindent
We follow \cite{tan2024fmri} in designing $G_{\text{FC}}$ and $G_{\text{SC}}$ architectures to preserve the symmetry structure of the connectomes. Altogether, the generators $G_{\text{FC}}$ and $G_{\text{SC}}$ and discriminators $D_{\text{FC}}$ and $D_{\text{SC}}$ are trained in a fashion following:

\begin{equation}
\begin{aligned}
\min _{G_{\text{FC}}, G_{\text{SC}}} \max _{D_{\mathrm{FC}}, D_{\mathrm{SC}}} \mathcal{L}_{obj} (G_{\text{FC}}, G_{\text{SC}}, D_{\text{FC}}, D_{\text{SC}})
\end{aligned}
\end{equation}

\noindent where the objective function of \textsf{SFC-GAN} follows:
\vspace{-0.1cm}
\begin{equation}
\begin{aligned}
\mathcal{L}_{obj}(G_{\text{FC}}, G_{\text{SC}}, & D_{\text{FC}}, D_{\text{SC}})  = \mathcal{L}_{GAN}(G_{\text{FC}}, D_{\text{SC}}, {x}_{\text{FC}}, \tilde{x}_{\text{SC}}) \\ &+ \mathcal{L}_{GAN}(G_{\text{SC}}, D_{\text{FC}}, {x}_{\text{SC}}, \tilde{x}_{\text{FC}}) \\ &+ \mathcal{L}_{cyc}(G_{\text{FC}}, G_{\text{SC}}) + \mathcal{L}_{\text{id}}(G_{\text{FC}}, G_{\text{SC}}) \\ &+ \mathcal{L}_{SP}(G_{\text{FC}}, G_{\text{SC}})
\end{aligned}
\label{eq:id}
\end{equation}

\vspace{-0.1cm}
\section{Experimental Results}

\subsection{Data acquisition and pre-processing}
We conducted experiments using two datasets: the Alzheimer's Disease Neuroimaging Initiative (ADNI) and the DUMC-MDD dataset, collected at Duke University Medical Center (DUMC). The ADNI dataset included 40 normal controls (CN), 40 individuals with mild cognitive impairment (MCI), and 40 Alzheimer’s disease (AD) patients. The fMRI data from ADNI were processed using the Data Processing Assistant for Resting-State fMRI (DPARSF) following \cite{Yan2019, Yan2010}. DMRI data were processed using the Pipeline for Analyzing braiN Diffusion imAges (PANDA) software \cite{cui2013panda} with a standard pipeline and deterministic tractography. The DUMC-MDD dataset consisted of 43 subjects, including 23 CN and 20 major depressive disorder (MDD) participants. Detailed pre-processing procedures for the DUMC-MDD dataset can be found in \cite{albert2019brain,wang2020}. For all FC and SC, we employed the Automated Anatomical Labeling (AAL) atlas to parcellate the brain into 116 regions of interest.

\begin{table*}[!t]

\caption{Classification performance on ADNI and DUMC-MDD datasets using original and translated SC and FC.} 
\label{tab:classification}

\resizebox{\textwidth}{!}{
\begin{tabular}{>{\centering\arraybackslash}p{3cm}|>{\centering\arraybackslash}p{3cm}|>{\centering\arraybackslash}p{4cm}|>{\centering\arraybackslash}p{2cm}>{\centering\arraybackslash}p{2cm}>{\centering\arraybackslash}p{2cm}>{\centering\arraybackslash}p{2cm}>{\centering\arraybackslash}p{2cm}}
\hline
{\centering \textbf{Classifier}}& {\centering \textbf{Dataset}} &{\centering \textbf{Testing data}} & {\centering \textbf{Accuracy}}	& {\centering \textbf{Precision}} & {\centering \textbf{Recall}} & {\centering \textbf{F1-Score}} & {\centering \textbf{AUC}}  \\
\hline

\multirow{12}{2cm}{\centering SVM} & \multirow{6}{2cm}{\centering ADNI} 
&  Real FC & 66.67 & 75.69 & 66.67 & 62.38 & 59.17 \\
& & Real SC & 41.67 & 45.83 & 41.67 & 41.67 & 61.11\\
& & Real FC and SC & 50.00 & 27.27 & 50.00 & 35.29 & 57.45 \\
&  & Translated FC & 16.67 & 11.11 & 16.67 & 12.50 & 52.18 \\
& & Translated SC & 41.67 & 27.78 & 41.67 & 33.33 & 57.78 \\
& & Translated FC and SC & 41.67 & 25.00 & 41.67 & 31.25 & 83.24 \\
\cline{2-8}
& \multirow{6}{2cm}{\centering DUMC-MDD} 
 &  Real FC & 69.23 & 47.93 & 69.23 & 56.64 & 50.00 \\
& & Real SC & 53.85 & 44.06 & 53.85 & 48.46 & 100.00 \\
& & Real FC and SC & 69.23 & 47.93 & 69.23 & 56.64 & 72.22 \\
& & Translated FC &  69.23 & 47.93 & 69.23 & 56.64 & 50.00 \\
& & Translated SC & 23.08 & 7.69 & 23.08 & 11.54 & 62.50 \\
& & Translated FC and SC & 69.23 & 47.93 & 69.23 & 56.64 & 72.22 \\

\hline
\hline

\end{tabular}

}
\end{table*}

\begin{figure*}[!htp]
\vspace{-0.2cm}
\captionsetup{skip=0pt}
\includegraphics[width=\textwidth]{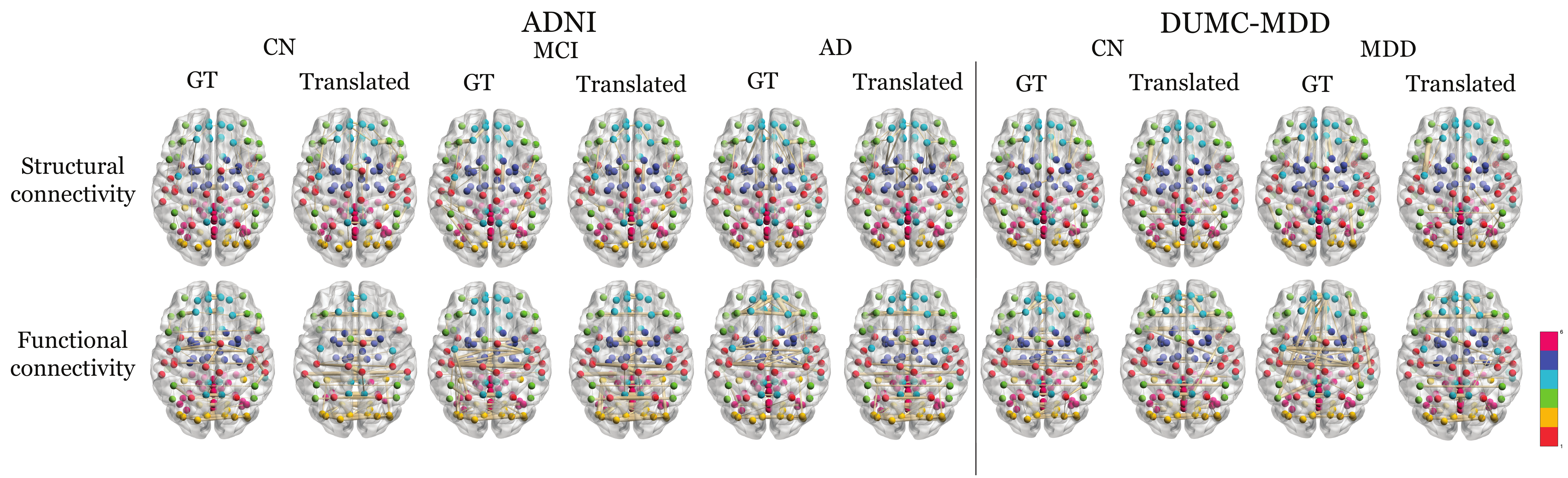}
\caption{Reconstruction results on both ADNI and DUMC-MDD datasets with top 5\% strongest connectivity in the brain space.
}
\label{fig:brainnet_recon}
\vspace{-0.5cm}
\end{figure*}

\subsection{Translated connectome evaluation}
 The performance of the translated connectomes was assessed using several metrics, including Mean Square Error (MSE), Mean Absolute Error (MAE), Structural Similarity Index Measure (SSIM), Pearson correlation, and cosine similarity. Additionally, we examined key graph properties, such as density, characteristic path length (CPL), global efficiency, and modularity, by calculating their Absolute Percentage Difference (APD) relative to the ground truth. We also examine the similarity performance with/without the incorporation of $\mathcal{L}_{SP}$. A classification study was conducted using a support vector machine (SVM) \cite{chang2011libsvm} to determine how well the translated connectomes preserved classification accuracy when compared to the ground truth.

\subsection{{Implementation details}} 
The \textsf{SFC-GAN} model was implemented in TensorFlow \cite{abadi2016tensorflow} The detailed architecture of \textsf{SFC-GAN} is illustrated in Fig.~\ref{fig:overview}c. \textsf{SFC-GAN} was trained for 200 epochs using the Adam optimizer \cite{kingma2014adam}. Both the learning rate and weight decay were set to $0.0001$. The model's performance was evaluated on two datasets: the ADNI dataset, with 96 subjects for training and 24 for testing, and the DUMC-MDD dataset, with 30 subjects for training and 13 for testing. For SVM classification, we measured accuracy, precision, recall, F1-score, and Area Under the Curve (AUC) on both real connectomes and those translated from another modality. The data partitioning scheme followed the same approach as used in training the proposed \textsf{SFC-GAN}, where the testing set of SC is translated to FC and vice versa for classification experiments.

\noindent{\textbf{Results and discussion:}} 
Figures \ref{fig:matrix_recon} and \ref{fig:brainnet_recon} illustrate the qualitative comparison between the translated connectomes and the ground truth, with matrix reconstruction outcomes and the brain space visualization of the top 5\% strongest connections after proportional thresholding for a representative subject. The analysis demonstrates that the translated FC and SC successfully preserve the topological structure of the connectomes, with the visual patterns  closely aligning with the ground truth. Furthermore, the top 5\% connectivity between the ground truth and the translated connectomes shows a high degree of similarity, indicating that \textsf{SFC-GAN} effectively captures the complex relationships between individual SC and FC, enabling accurate connectomes translation.

Table \ref{tab:results_similarity} presents the similarity metrics and graph property evaluations for the translated connectomes compared to the ground truth. The results demonstrate consistently lower MSE and MAE across both datasets, along with highly comparable graph properties in the translated FC for both testbeds. Notably, the incorporation of the structure-preserving loss, $\mathcal{L}_{SP}$, enhances performance across all metrics, underscoring its effectiveness in training \textsf{SFC-GAN} to maintain the connectomes topological structure. The translated FC connectomes, across both the ADNI and DUMC-MDD datasets, exhibit better graph property alignment with the ground truth and higher cosine similarity, although they show lower SSIM and Pearson correlation compared to the translated SC connectomes. This suggests that while the generated FC from SC preserves the overall topological structure, it may not capture the finer details of SC. On the other hand, the translated SC connectomes display consistent performance across both similarity and graph property evaluations, indicating a stronger ability of the model to capture the relationship from FC to SC than from SC to FC. Table \ref{tab:classification} reports the classification performance on the ADNI and DUMC-MDD datasets. The results show that the translated SC and the combined SC and FC connectomes in the ADNI testbed, as well as the translated FC and the combined SC and FC connectomes in the DUMC-MDD testbed, achieve classification performance that is closely aligned with the ground truth. This suggests that the translated connectomes are well-suited for subsequent analyses. However, we observed that the translated FC in the ADNI testbed and the translated SC in the DUMC-MDD testbed exhibited inferior classification performance, indicating areas where further refinement is needed to enhance model performance, offering a powerful approach to understanding and modeling the complex dynamics of brain connectivity.

\section{Conclusion}

We developed a novel \textsf{SFC-GAN}, incorporating specialized layer and a structure-preserving loss, to enable both direct and inverse mapping between FC and SC while maintaining the topological order and symmetry properties. Our qualitative and quantitative results demonstrate that the model can accurately translate between SC and FC. 


\label{sec:refs}

\newpage

\bibliographystyle{IEEEbib}
{\bibliography{refs}}

\end{document}